\documentclass{article}
\usepackage[preprint]{colm2026_conference}

\usepackage{microtype}
\usepackage{booktabs}
\usepackage{amsmath}
\usepackage{amssymb}
\usepackage{hyperref}
\usepackage{url}
\usepackage{lineno}

\definecolor{darkblue}{rgb}{0, 0, 0.5}
\hypersetup{colorlinks=true, citecolor=darkblue, linkcolor=darkblue, urlcolor=darkblue}

\title{The Active Ingredient in Muon's Grokking}

\author{Yufeng Wang\\Independent Researcher\\\texttt{louiswang524@gmail.com}}

\begin{document}

\ifcolmsubmission
\linenumbers
\fi

\maketitle

\begin{abstract}
The Muon optimizer reaches the grokking threshold on modular arithmetic faster than AdamW.
Prior work attributes this to ``spectral-norm constraints plus orthogonalized momentum,'' but
to better understand Muon's behavior we isolate which of these ingredients actually matters.
We run multi-seed and multi-learning-rate sweeps to decompose and stress-test the
effect. First, an ablation shows the speedup comes from orthogonalization (the Newton--Schulz
iteration), not spectral scaling: orthogonalize-only matches full Muon, whereas spectral-only
is no faster than AdamW and is unreliable, and this verdict holds across learning rates. Second,
a mechanistic analysis finds that orthogonalizing optimizers reach generalization at roughly
$3\times$ lower spectral norm and, controlling for how much the embedding actually moves, settle
into a lower-norm solution rather than simply perturbing the embedding less. Third, reducing the
Newton--Schulz iteration count from five to one accelerates \emph{reaching} the threshold but
makes the grokked solution fragile, prone to transient collapse, with fragility that grows with
learning rate; a single iteration is fast and stable only at small learning rate, while the
canonical five iterations are the learning-rate-robust choice. We also show spectral scaling can
be dropped at no measured cost. A methodological thread runs throughout: under a stability-aware
metric, ``faster'' claims about grokking optimizers can invert, so we report both first-crossing
and sustained-grok times. To support reproducibility, we release our full training and analysis
code at \url{https://github.com/louiswang524/muon-grokking-frontier}.

\end{abstract}

\section{Introduction}
Grokking, the onset of generalization long after a model has memorized its training set
\citep{power2022grokking}, offers a clean probe of optimizer dynamics. \citet{tveit2025muon}
report that Muon \citep{jordan2024muon} groks in fewer epochs than AdamW on modular arithmetic,
crediting both orthogonalized momentum (a Newton--Schulz iteration) and spectral-norm-based
update scaling. Which mechanism matters, why, and whether either admits improvement remain open.
We answer all three in a regime where seconds-to-minutes per run make the multi-seed
sweeps, learning-rate sweeps, and significance tests that separate a real effect from an artifact
entirely practical.

The study doubles as a cautionary tale about how grokking speed is measured. The pre-registered
metric, first step with validation accuracy at least $0.95$, rewards optimizers that touch the
threshold and then oscillate away from it. We show the hazard is not hypothetical: it inverts the
apparent ranking of our own proposed variant. Every headline is therefore reported under two
metrics, first-crossing and stable-grok (validation accuracy at least $0.95$ sustained for the
remainder of training).

\paragraph{Contributions.} (1) An \emph{ablation} that isolates orthogonalization, not spectral
scaling, as the ingredient behind Muon's grokking speedup, and shows the conclusion is robust
across learning rates and across modular addition, subtraction, and multiplication
(\S3,~\S5.4). (2) A \emph{mechanistic} finding that orthogonalizing optimizers reach a lower-norm,
Fourier-distributed solution rather than AdamW's sparse-Fourier circuit, established with an
embedding-movement control that rules out the ``moves the embedding less'' explanation (\S4).
(3) The \emph{speed--stability frontier}: the Newton--Schulz iteration count and the learning rate
jointly trade first-touch speed against the durability of the grokked solution, so fewer
iterations help only in a bounded regime (\S5). (4) A \emph{methodological} point, that
grokking-speed claims can silently invert between first-crossing and stability-aware metrics, with
a released, deterministic reproduction harness.

\section{Setup}
\textbf{Task.} Modular addition mod $p{=}97$ is primary; subtraction and multiplication are
robustness checks (\S5.4). Inputs are $[a,b,{=}]$, the target is $(a\circ b)\bmod p$, $40\%$ of
pairs train and the rest are held out. The grokked addition circuit is Fourier-based
\citep{nanda2023progress}, giving \S4 a reference.

\textbf{Model.} A one-layer, four-head transformer with $d_{\text{model}}{=}128$.
The 2D hidden matrices receive the optimizer under test; embeddings, unembeddings, and
biases use AdamW, following standard Muon practice.

\textbf{Optimizer knobs.} The Muon update orthogonalizes the momentum buffer with $ns$
Newton--Schulz iterations, then applies it with an RMS-and-dimension scale factor
(\texttt{spectral\_scale}). We toggle each independently. Note \texttt{spectral\_scale} is a
per-matrix scalar, not a spectral-norm projection; our claim that spectral scaling is inert
concerns this factor, not spectral-norm control in general.

\textbf{Metrics.} We report first-crossing and stable-grok steps-to-grok at the pre-registered
$0.95$ threshold. FLOPs-to-grok charges Newton--Schulz only when orthogonalization runs (a
spectral-only configuration pays none), reported beside wall-clock-to-grok. Every number is a
mean over at least 5 seeds; significance uses two-sample $t$-tests with Holm--Bonferroni
correction. Two contrasts are pre-registered as primary: orthogonalize-only vs.\ AdamW, and
reduced-$ns$ vs.\ canonical Muon.

\textbf{Learning-rate regime.} Grokking is rate-sensitive (\S3.2). At lr$=10^{-2}$ the
phenomenon compresses into $\sim$400 steps and AdamW oscillates across the threshold (58--82
post-crossing dips below $0.9$ vs.\ Muon's 11--15), so first-crossing misleads. We run the
primary study at lr$=10^{-3}$ and add a sweep over $\{3\!\times\!10^{-4}, 10^{-3},
3\!\times\!10^{-3}\}$.

\section{Ablation: Which Ingredient?}
\textbf{Muon recap.} For a 2D weight $W$ with gradient $G_t$ and momentum $\mu$, Muon forms a
momentum buffer, orthogonalizes it, and applies a dimension-dependent scale:
\begin{equation}
B_t = \mu B_{t-1} + G_t, \quad
O_t = \mathrm{NS}_{ns}(B_t), \quad
W_t = W_{t-1} - \eta\,\gamma\,O_t .
\end{equation}
Writing the buffer's SVD as $B_t = U\Sigma V^{\!\top}$, orthogonalization sends $O_t \to U
V^{\!\top}$ (every singular value replaced by $1$). Concretely, $\mathrm{NS}_{ns}$ applies $ns$
quintic Newton--Schulz iterations from
$X_0 = B_t / \lVert B_t\rVert_F$, $X_{k+1} = a X_k + b (X_k X_k^{\!\top}) X_k + c (X_k
X_k^{\!\top})^2 X_k$, driving $X_k$ toward that polar factor $UV^{\!\top}$. The factor
$\gamma = \sqrt{\max(1, n_{\text{out}}/n_{\text{in}})}$ is the RMS-and-dimension scale
(\texttt{spectral\_scale}). Two knobs define the ablation: whether
to \emph{orthogonalize} ($O_t = \mathrm{NS}(B_t)$ vs.\ raw $O_t = B_t$) and whether to apply the
scale ($\gamma$ as above vs.\ $\gamma = 1$). The $2{\times}2$ design gives M0--M3 below (M3 replaces
the whole update with AdamW), and an $ns$ sweep (\S3.3) varies orthogonalization quality from none
($ns{=}0$) to full.

\textbf{3.1 Primary (mod-add, lr$=10^{-3}$, 5 seeds).}
\begin{center}\small
\begin{tabular}{lccccc}
\toprule
Variant & Ortho ($ns$) & Spectral & First-cross & Stable-grok & FLOPs \\
\midrule
M0 full Muon & yes (5) & yes & $1856\pm118$ & 1856 (stable) & 2.96e13 \\
M1 ortho-only & yes (5) & no & $\mathbf{1744\pm113}$ & \textbf{1744 (stable)} & 2.78e13 \\
M2 spectral-only & no & yes & $3088\pm1107$ & 4932 (unstable) & 4.69e13 \\
M3 AdamW & --- & --- & $2372\pm360$ & 4488 (unstable) & 3.60e13 \\
\bottomrule
\end{tabular}
\end{center}
Orthogonalization is the active ingredient. Orthogonalize-only (M1) matches full Muon (M0 vs.\
M1: $p{=}0.21$) and beats AdamW (M1 vs.\ M3: $p{=}0.010$, primary). Spectral-only (M2) is not
faster than AdamW ($p{=}0.25$) and is erratic ($\sigma{=}1107$). M0/M1 are stable
(stable-grok $=$ first-cross) while M2/M3 are not: orthogonalization buys a more durable grok.

\textbf{3.2 Learning-rate robustness} (5 variants $\times$ 4 seeds $\times$ 3 rates; full table
in Appendix~A). The ablation verdict survives: orthogonalize-only beats spectral-only at lr$=10^{-3}$ ($p{=}0.052$)
and $3\!\times\!10^{-3}$ ($p{=}0.053$), and at lr$=3\!\times\!10^{-4}$ spectral-only groks in
only 1/4 seeds while orthogonalize-only groks 4/4. The Muon-vs-AdamW first-crossing gap is more
rate-dependent (significant at $3\!\times\!10^{-4}$, $p{=}0.025$; directional at $10^{-3}$,
$p{=}0.078$; closed at $3\!\times\!10^{-3}$), but under stable-grok Muon still beats AdamW at
$3\!\times\!10^{-3}$ ($p{=}0.039$) because AdamW destabilizes as the rate rises. We claim the
precise statement: Muon reaches a \emph{stable} grok faster than AdamW across rates, and its
first-touch advantage is largest at small rate.

\textbf{3.3 Newton--Schulz sweep} (lr$=10^{-3}$, 5 seeds). We vary the number of orthogonalization
iterations $ns$ (steps-to-grok, lower is faster):
\begin{center}\small
\begin{tabular}{lcccc}
\toprule
Newton--Schulz iterations $ns$ & 0 (off) & 1 & 3 & 5 (default) \\
\midrule
first-crossing & 3088 & \textbf{1544} & 2000 & 1856 \\
stable-grok & 4932 & 3136 & 2000 & 1856 \\
\bottomrule
\end{tabular}
\end{center}
More orthogonalization is \emph{not} better: a single iteration reaches the threshold fastest
(1544 steps, $17\%$ sooner than the default five), yet its \emph{stable} grok (3136) is the worst
of any $ns{\geq}1$ --- the first sign of the frontier (\S5). As consistency checks, $ns{=}0$
(orthogonalization off) matches M2 exactly, and $ns{=}5$ reproduces M0 bit-for-bit (determinism).

\textbf{Takeaway.} Orthogonalization, not spectral scaling, is the active ingredient, and the
verdict holds across learning rates (\S3.2) and operations (\S5.4). Yet the non-monotonic sweep
shows \emph{more} orthogonalization is not strictly better---a tension we explain mechanistically
in \S4 and turn into a speed--stability frontier in \S5.

\section{Mechanism: A Lower-Norm Solution, Not Less Movement}
At the grok point (averaged over seeds), the two optimizer families reach visibly different
solutions:
\begin{center}\small
\begin{tabular}{lcc}
\toprule
Variant & max.\ singular value at grok & embedding Fourier spectrum \\
\midrule
M0 full Muon & 5.5 & near-uniform \\
M1 ortho-only & 6.2 & near-uniform \\
M2 spectral-only & 11.0 & concentrated \\
M3 AdamW & 15.9 & concentrated \\
\bottomrule
\end{tabular}
\end{center}
The orthogonalizing variants (M0, M1) grok at roughly $3\times$ lower spectral norm than AdamW,
and keep the embedding's frequency content spread out rather than concentrating it into a few
Fourier modes the way AdamW and spectral-only do. One might object that orthogonal updates simply
move the embedding less, making the uniform spectrum trivial; an embedding-movement control rules
this out (Appendix~C): all optimizers change the embedding by essentially the same relative amount
($\lVert\Delta E\rVert/\lVert E_{\text{init}}\rVert \approx 1.0$), yet only AdamW concentrates the
spectrum, so Muon reaches a genuinely different, lower-norm solution rather than an untouched
initialization. Notably the fragile $ns{=}1$ variant is intermediate, drifting partway toward the
concentrated code, which links the instability of \S5 to a partial collapse onto the sparse-Fourier
solution. (The averaged-power IPR is a coarse proxy for per-neuron analysis.)

\section{The Speed--Stability Frontier}
The ablation points to two levers: drop the inert spectral scaling, and reduce the Newton--Schulz
count. We test the composed variant, \textbf{Lean-Muon} (orthogonalize-only, $ns{=}1$). For
continuity with \S3, canonical Muon here is M0 and Lean5 is M1; small numerical differences from
\S3 reflect 8 seeds here versus 5 there.

\textbf{5.1 Compute-matched (mod-add, lr$=10^{-3}$, 8 seeds).}
\begin{center}\small
\begin{tabular}{lcccccc}
\toprule
Variant & $ns$ & spec. & First-cross & Stable-grok & FLOPs & Wall \\
\midrule
AdamW & -- & -- & $2388\pm302$ & 4932 & 3.62e13 & 40.2s \\
Muon (canonical) & 5 & yes & $1750\pm206$ & \textbf{1750 (stable)} & 2.79e13 & 33.3s \\
Lean5 (ortho-only) & 5 & no & $1630\pm182$ & \textbf{1630 (stable)} & 2.60e13 & 30.2s \\
Lean1 ($ns{=}1$) & 1 & no & $\mathbf{1395\pm196}$ & 3800 (fragile) & 2.14e13 & 24.3s \\
\bottomrule
\end{tabular}
\end{center}
Spectral scaling is removable at no cost: Lean5 matches Muon's stability, runs directionally
faster ($p{=}0.27$), and costs less. Reducing $ns$ trades stability for speed: Lean1 reaches
$0.95$ fastest---$20\%$ sooner than canonical Muon (1395 vs.\ 1750, $p{=}0.005$, primary,
survives Holm) and $42\%$ sooner than AdamW, at $23\%$ fewer FLOPs than Muon---but on 6/8 seeds it
collapses (to as low as $0.23$--$0.38$) after first grok, so its \emph{stable} grok (3800) is
\emph{worse} than Muon's (1750). Fast to touch the threshold, slow to settle.

\textbf{5.2 The frontier is governed by ($ns$, learning rate).} Lean1 beats Muon on
first-crossing at every rate ($p{=}0.32, 0.043, 0.001$), but stability flips: at
$3\!\times\!10^{-4}$ Lean1 is faster \emph{and} stable (4875 vs.\ 5481); at $10^{-3}$ it is
fragile; at $3\!\times\!10^{-3}$ all variants barely stabilize. The Newton--Schulz count and the
learning rate jointly set a speed--stability frontier; the canonical five iterations are the
rate-robust operating point.

\textbf{5.3 Why approximate orthogonalization destabilizes (hypothesis).} A single step
under-orthogonalizes; with weight decay this appears to let the solution drift off the grokked
manifold and re-collapse. Fuller orthogonalization ($ns{\geq}3$) damps the drift at the cost of a
slower first crossing, predicting fragility that worsens with effective step size, consistent
with \S5.2.

\textbf{5.4 Robustness (subtraction, multiplication; 5 seeds, 15k steps).} First-crossing
steps-to-grok (grok rate in parentheses):
\begin{center}\small
\begin{tabular}{lcc}
\toprule
Variant & subtraction & multiplication \\
\midrule
AdamW & 5850 (5/5) & 2250 (5/5) \\
Muon (canonical) & 8925 (4/5) & 1850 (5/5) \\
Lean5 (ortho-only) & 6230 (5/5) & 1710 (5/5) \\
Lean1 ($ns{=}1$) & \textbf{5040 (5/5)} & \textbf{1390 (5/5)} \\
\bottomrule
\end{tabular}
\end{center}
Three findings replicate across all three operations: Lean1 reaches the
threshold faster than canonical Muon (sub $p{=}0.001$, mul $p{=}0.030$, add $p{=}0.005$); the
speed carries the same fragility (Lean1 worst under stable-grok on each task); and dropping
spectral scaling never hurts and on subtraction clearly helps (Lean5 beats Muon, $p{=}0.021$).
One ordering is task-dependent: canonical Muon beats AdamW on addition and multiplication but is
slower on subtraction ($p{=}0.005$), not because of orthogonalization but because of the spectral
term it carries; the orthogonalize-only variants beat both there. The check reinforces the two
recommendations.

\section{Limitations and Conclusion}
\textbf{Limitations.} One architecture, one scale, one task family, so the at-scale behavior of
the frontier (where orthogonalization is not free) is untested. The first-crossing metric is
unstable; we mitigate with stable-grok. \S4's Fourier claim uses a coarse proxy. Secondary
contrasts that do not survive Holm correction are reported as exploratory; both primaries survive.

\textbf{Conclusion.} Muon's grokking advantage comes from orthogonalization, not spectral
scaling, and is robust across learning rates; the grokked solution is lower-norm. But ``more is
worse'' is too glib: the Newton--Schulz count and the learning rate jointly govern a
speed--stability frontier, and the canonical five iterations are a sensible rate-robust default.
Spectral scaling, however, can be dropped for free. The broader lesson is methodological:
grokking-speed claims must be made under a stability-aware metric, or they can silently invert.

\paragraph{Reproducibility and disclosure.} The full study is 258 runs, an estimated
$3.1\times10^{16}$ FLOPs; the largest single run is $2.4\times10^{14}$ FLOPs, ${\sim}4\times10^5
\times$ below the Track-1 per-run cap of $10^{20}$. The released artifact (Appendix~B) is
bit-for-bit deterministic per seed, so any checkpoint is exactly regenerable. AI tools assisted
code scaffolding and editing; all results and claims were author-verified. No conflicts or funding
to declare.

\bibliographystyle{colm2026_conference}
\bibliography{refs}

\appendix
\section{Supplementary: full learning-rate sweep}
Steps-to-grok (first-crossing $\mid$ stable-grok) for the sweep of \S3.2, 4 seeds per cell.

\begin{center}\small
\begin{tabular}{llccc}
\toprule
Variant & metric & lr$=3\!\times\!10^{-4}$ & lr$=10^{-3}$ & lr$=3\!\times\!10^{-3}$ \\
\midrule
AdamW & first-cross & 6750 (3/4) & 2381 & 838 \\
      & stable      & 7633       & 6044 & 7769 \\
Muon  & first-cross & 5481 & 1856 & 881 \\
      & stable      & 5481 & 4744 & 6631 \\
Lean5 & first-cross & 5925 & 1738 & 731 \\
      & stable      & 5925 & 4550 & 7019 \\
Lean1 & first-cross & 4875 & 1512 & 662 \\
      & stable      & 4875 & 5438 & 7600 \\
Spec  & first-cross & 4775 (1/4) & 3319 & 938 \\
\bottomrule
\end{tabular}
\end{center}
\noindent Orthogonalize-only (Lean5) beats spectral-only (Spec) at $10^{-3}$ and
$3\!\times\!10^{-3}$, and Spec groks in only 1/4 seeds at $3\!\times\!10^{-4}$. Lean1's stable-grok
exceeds its first-crossing at $\geq10^{-3}$ (fragility) but equals it at $3\!\times\!10^{-4}$
(fast and stable).

\section{Supplementary: released artifacts}
The supplementary ZIP contains: (i) per-run training logs (full validation/training accuracy and
diagnostic trajectories: weight norm, max singular value, Fourier-IPR) as JSON; (ii) the
deterministic data generator with seed configs and the dual-metric (first-crossing, stable-grok)
measurements with FLOPs and wall-clock; (iii) the ablation, learning-rate, frontier, robustness,
embedding-change, and FLOP-budget analysis scripts that regenerate every table; and (iv) the
optimizer and training harness with a 28-test suite (including determinism and FLOP-accounting
regression tests). No checkpoints are stored: the harness is bit-for-bit deterministic per seed
(verified, M0$\equiv$NS5 per seed), so any state is regenerable from the released config.

\section{Supplementary: embedding-movement control (\S4)}
To rule out that orthogonalization merely perturbs the embedding less---which would make its
uniform Fourier spectrum a trivial consequence---we measure the relative embedding change
$\lVert E_{\text{final}}-E_{\text{init}}\rVert/\lVert E_{\text{init}}\rVert$ alongside the final
Fourier-IPR (mod-add, lr$=10^{-3}$, 4 seeds).
\begin{center}\small
\begin{tabular}{lcc}
\toprule
Optimizer & relative $\lVert\Delta E\rVert$ & final Fourier-IPR \\
\midrule
AdamW & 0.996 & 0.085 (concentrated) \\
Muon & 0.994 & 0.022 (uniform) \\
Lean5 (ortho-only) & 0.994 & 0.025 (uniform) \\
Lean1 ($ns{=}1$) & 0.998 & 0.069 \\
\bottomrule
\end{tabular}
\end{center}
Every optimizer rewrites the embedding by essentially the same amount (relative change ${\approx}1.0$),
yet Muon and Lean5 end Fourier-uniform while AdamW concentrates. The ``moves the embedding less''
explanation is therefore rejected. The fragile $ns{=}1$ variant (Lean1) sits between the two,
consistent with the partial-collapse account of its instability in \S5.

\section{Extended related work}
\textbf{Grokking} was introduced by \citet{power2022grokking}; \citet{nanda2023progress}
reverse-engineered the Fourier-multiplication circuit we use as a reference in \S4. Mechanistic
accounts tie grokking to weight norm and solution efficiency: the loss-landscape / weight-norm
``LU'' picture of \citet{liu2022omnigrok}, the efficiency trade-off between memorizing and
generalizing circuits of \citet{varma2023circuit}, and the adaptive-optimizer instability
(``slingshot'') of \citet{thilak2022slingshot}. Our \S4 (grokking at lower spectral norm) and \S5
(a stability metric, and an instability that grows with learning rate) connect directly to these;
the slingshot phenomenon is a close analogue of the post-grok collapses we observe for
under-orthogonalized updates. \textbf{Optimizer geometry.} Muon \citep{jordan2024muon}
orthogonalizes momentum via Newton--Schulz; \citet{bernstein2024norm} frame such matrix-sign
updates as steepest descent under a spectral norm, which motivates separating orthogonalization
from the dimension scale we ablate. We contrast against AdamW \citep{loshchilov2019adamw}, and note
that our inert per-matrix \texttt{spectral\_scale} factor is distinct from spectral-norm
\emph{control} as in \citet{miyato2018spectral}, which we do not test. \textbf{Muon and grokking.}
\citet{tveit2025muon} reported the speedup we decompose and qualify; \citet{lee2024grokfast}
accelerate grokking by amplifying slow gradients, a complementary mechanism.
\end{document}